\pdfoutput=1

\documentclass[11pt]{article}

\usepackage{acl}

\usepackage{times}
\usepackage{latexsym}

\usepackage[T1]{fontenc}

\usepackage[utf8]{inputenc}

\usepackage{microtype}

\usepackage{inconsolata}
\usepackage{amssymb}
\usepackage{amsmath}
\usepackage{multirow}
\usepackage{multicol}
\usepackage{graphicx} 
\usepackage{booktabs}
\usepackage{enumitem}
\usepackage{colortbl}  
\usepackage{xcolor}
\usepackage{array}

\usepackage{tcolorbox}
\newtcolorbox[list inside=prompt,auto 
counter,number within=section]{prompt}[1][]{
    colbacktitle=black!60,
    coltitle=white,
    fontupper=\footnotesize,
    boxsep=5pt,
    left=0pt,
    right=0pt,
    top=0pt,
    bottom=0pt,
    boxrule=1pt,
    title={#1},
    #1, 
}
\title{End-to-End Beam Retrieval for Multi-Hop Question Answering}

\author{
  Jiahao Zhang\textsuperscript{\rm \dag} \quad
  Haiyang Zhang\textsuperscript{\rm \dag}\thanks{Corresponding author} \quad
  Dongmei Zhang\textsuperscript{\rm \dag} \quad
  Yong Liu\textsuperscript{\rm \ddag}\quad
  Shen Huang\textsuperscript{\rm \ddag}\\
  \\
  \textsuperscript{\rm \dag}Beijing University of Posts and Telecommunications, Beijing, China \\
  \small \textsuperscript{}{\{relyourself, zhhy, zhangdm\}@bupt.edu.cn} \\
  \\
  \textsuperscript{\rm \ddag}Tencent Research, Beijing, China \\
  \small \textsuperscript{}{\{owenyongliu, springhuang\}@tencent.com}
}

\begin{document}
\maketitle
\begin{abstract}
Multi-hop question answering (QA) involves finding multiple relevant passages and step-by-step reasoning to answer complex questions, indicating a retrieve-and-read paradigm. However, previous retrievers were customized for two-hop questions, and most of them were trained separately across different hops, resulting in a lack of supervision over the entire multi-hop retrieval process and leading to poor performance in complicated scenarios beyond two hops. In this work, we introduce Beam Retrieval, an end-to-end beam retrieval framework for multi-hop QA. This approach models the multi-hop retrieval process in an end-to-end manner by jointly optimizing an encoder and two classification heads across all hops. 
Moreover, Beam Retrieval maintains multiple partial hypotheses of relevant passages at each step, expanding the search space and reducing the risk of missing relevant passages. To establish a complete QA system, we incorporate a supervised reader or a large language model (LLM).  Experimental results demonstrate that Beam Retrieval achieves a nearly 50\% improvement compared with baselines on challenging MuSiQue-Ans, and it also surpasses all previous retrievers on HotpotQA and achieves 99.9\% precision on 2WikiMultiHopQA. Providing high-quality context, Beam Retrieval helps our supervised reader achieve new state-of-the-art performance and substantially improves the few-shot 
 QA performance of LLMs\footnote{Code is available at \url{https://github.com/canghongjian/beam_retriever}}.
\end{abstract}

\section{Introduction}
\begin{figure}[!t]
  \centering
  \includegraphics[width=\linewidth]{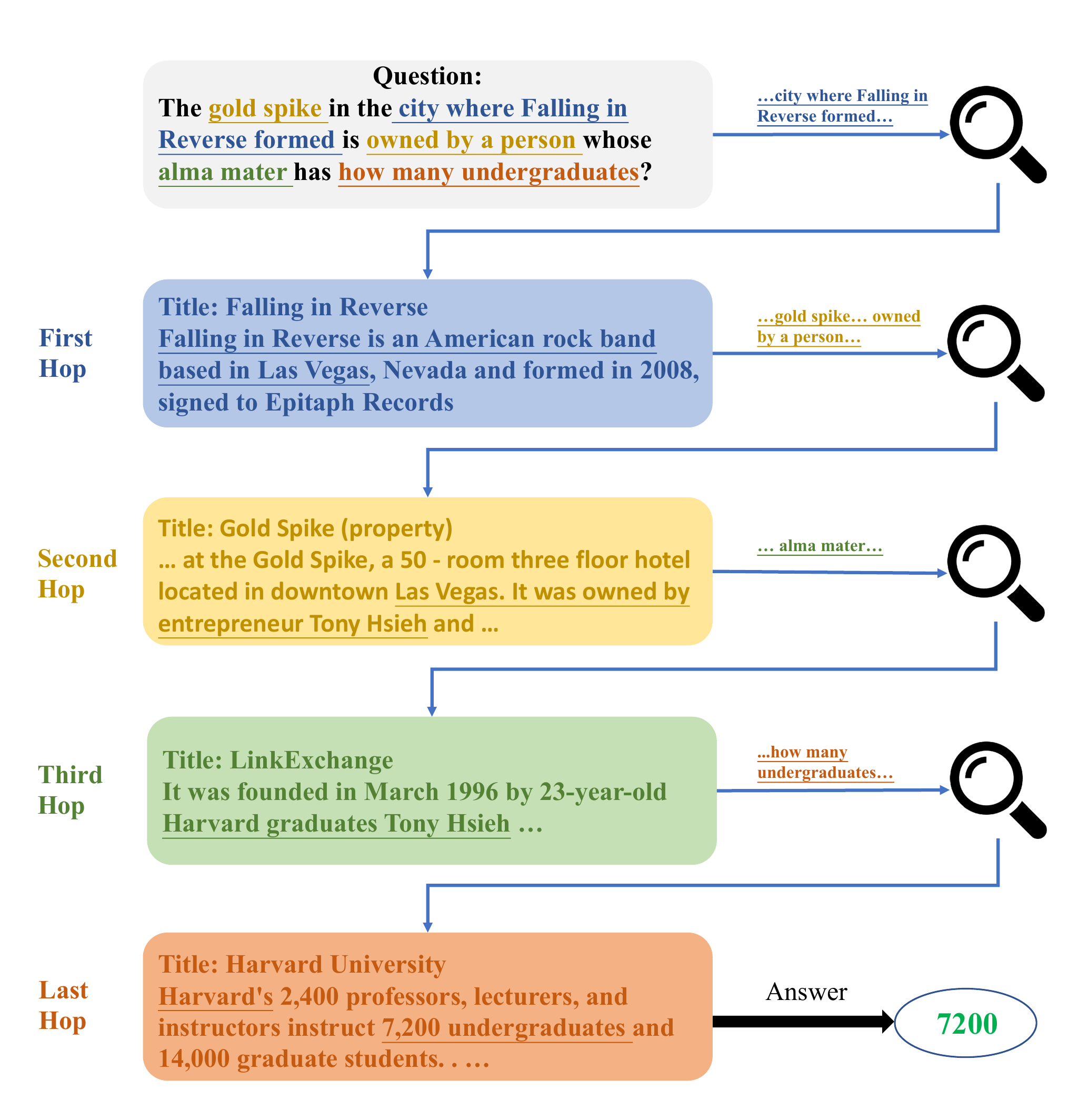}
  \caption{An example of multi-hop QA from MuSiQue-Ans benchmark. This complicated 4-hop question requires the model to select relevant passages based on the question and previously chosen passages.}
  \label{fig:examples}
\end{figure}
Question Answering (QA) has been a mainstream research in natural language processing (NLP) for a long time. With the development of pretrained language models (PLMs), simple QA tasks can be solved by adopting a BERT-like PLM \cite{devlin-etal-2019-bert}. As a result, researchers have been increasingly drawn to more complex QA benchmarks, such as multi-hop QA. This presents a significant challenge, as it requires reasoning across multiple and diverse passages to accurately answer complicated multi-hop questions. Many high-quality multi-hop QA datasets have been introduced, such as HotpotQA \cite{yang-etal-2018-hotpotqa}, 2WikiMultiHopQA \cite{DBLP:journals/corr/abs-2011-01060}, MuSiQue \cite{musique} and so on. Figure~\ref{fig:examples} illustrates an example of an actual question taken from MuSiQue-Ans dataset.

Mainstream methods for multi-hop QA often follow a retrieve-and-read paradigm \cite{chen-etal-2017-reading, retrieving_and_reading}, including a passage retriever to filter out extraneous information and a reader to obtain the final answer \cite{chen-etal-2017-reading, DBLP:conf/aaai/TuHW0HZ20, mdr, zhao-etal-2021-multi-step, DBLP:journals/corr/abs-2107-11823, musique, li2023easy, zhangyue-etal-2023-rethinking}. However, these methods have primarily focused on two-hop scenarios, exhibiting limited adaptability to more complex situations beyond two-hops. Additionally, while multi-hop retrieval requires identifying the next hop passage based on the question and previously selected passages (see figure~\ref{fig: examples}), few of them focus on supervision over the entire retrieval process. Furthermore, these retrievers exhibit limited robustness, as the entire retrieval process is susceptible to failure if the first stage identifies irrelevant passages. In conclusion, previous retrievers perform poorly when handling questions with more than 2 hops and provide low-quality context for downstream QA tasks.

To address the described problems, we propose Beam Retrieval, an end-to-end beam retrieval framework for multi-hop QA. Beam Retrieval utilizes an encoder and two classification heads to model the entire multi-hop retrieval process in an end-to-end manner and can be adapted to a question with a variable hop. During training, Beam Retrieval accumulates the loss at each step and jointly optimizes the encoder and two classification heads in the backpropagation phase, enabling the model to learn the entire retrieval process. During inference, Beam Retrieval searches the relevant passage at each step until the highest predicted score falls below a predefined threshold. In summary, Beam Retrieval produces a chain of relevant passages with the highest score using a single forward pass, effectively learning the entire multi-hop retrieval process. Moreover, we employ the beam search paradigm by keeping track of multiple partial hypotheses of relevant passages at each step. This approach enables our model to learn more negative passage pairs in the expanded search space, enhances the probability of obtaining the truly relevant passages, and mitigates the impact of retrieval errors that may occur in the early stages. To reduce the gap between training and reasoning, Beam Retrieval is designed to reason using the same beam size as it employs during training.

Beam Retrieval can also serve as a plugin in the QA domain, providing high-quality relevant context and enhancing the performance of downstream QA tasks. Based on Beam Retrieval, we implement a multi-hop QA system to extract the answers by incorporating a supervised reader \cite{li2023easy, zhangyue-etal-2023-rethinking} following conventional machine reading comprehension setting or a few-shot large language model (LLM) \cite{gpt3, gpt4}. We validate Beam Retrieval by extensive experiments on three benchmark datasets MuSiQue-Ans, HotpotQA and 2WikiMultihopQA, and experimental results demonstrate that Beam Retrieval surpasses all previous retrievers by a large margin. Consequently, Beam Retrieval substantially improves the QA performance of downstream QA readers on all three datasets.

We highlight our contributions as follows:
\begin{itemize}
\item We propose Beam Retrieval, which models the entire multi-hop retrieval process in an end-to-end manner by jointly optimizing an encoder and two classification heads across all hops. Designed to handle questions with variable hops, Beam Retrieval shows great performance, especially in complex scenarios beyond two hops.
\item Our Beam Retrieval keeps multiple hypotheses of relevant passages at each step during end-to-end training and inference, which mitigates the impact of retrieval errors that may occur in the early steps. This beam search paradigm brings further improvement.
\item We evaluate our multi-hop QA system on three multi-hop QA datasets to validate the effectiveness of Beam Retrieval. Beam Retrieval achieves a nearly 50\% improvement compared with baselines on challenging MuSiQue-Ans, and it also surpasses all previous retrievers on HotpotQA and achieves 99.9\% precision on 2WikiMultiHopQA. Providing high-quality context, Beam Retrieval helps our supervised reader achieve new state-of-the-art performance and substantially improves the few-shot QA performance of LLMs.
\end{itemize}

\section{Related Work}
\paragraph{Retrievers in Multi-Hop QA}
Mainstream methods for multi-hop QA often follow a retrieve-and-read paradigm \cite{chen-etal-2017-reading, retrieving_and_reading}, where a retriever is used to find passages relevant to the multi-hop question, followed by a reader that answers the question based on the retrieved content. Previous retrievers focus on two types of multi-hop QA settings: the open-domain setting and the reading comprehension setting. In the open-domain setting, models are required to retrieve relevant passages within a large-scale corpus, while the reading comprehension setting involves searching within a smaller set of candidate passages. In open-domain multi-hop QA, retrievers can be categorized into semantic retrieval methods like BM25 \cite{chen-etal-2017-reading} and dense retrieval methods like MDR \cite{mdr} and BeamDR \cite{zhao-etal-2021-multi-step}. Retrievers in the reading comprehension setting are almost cross-encoders, divided into two types. One type is the one-step methods. SAE \cite{DBLP:conf/aaai/TuHW0HZ20} and MuSiQue SA Selector \cite{musique} concatenate each candidate passage and the question as inputs fed to BERT, then select out the most relevant passages with the highest scores. Such methods do not utilize the dependency between relevant passages, resulting in a limited performance. The other type is the two-step method. S2G \cite{DBLP:journals/corr/abs-2107-11823} and FE2H \cite{li2023easy} select the first hop passage in the same way as one-step. In the second stage, they identify the second hop relevant passage by pairing the selected passage with the other candidate passages. R$^3$ \cite{zhangyue-etal-2023-rethinking} selects three passages in the first stage, then combines them two by two and identifies the true passage pair in the second stage. Notice that the unselected passages in the first stage will not be utilized in the second stage, leaving limitations in retrieval. 
The Beam Retrieval proposed in this paper, primarily aimed at the reading comprehension setting, similarly introduces the idea of beam search as in BeamDR. However, unlike BeamDR, Beam Retrieval emphasizes modeling the entire multi-hop retrieval process and dealing with complex scenarios beyond two hops.

\begin{figure*}[!ht]
  \centering
  \includegraphics[width=\linewidth]{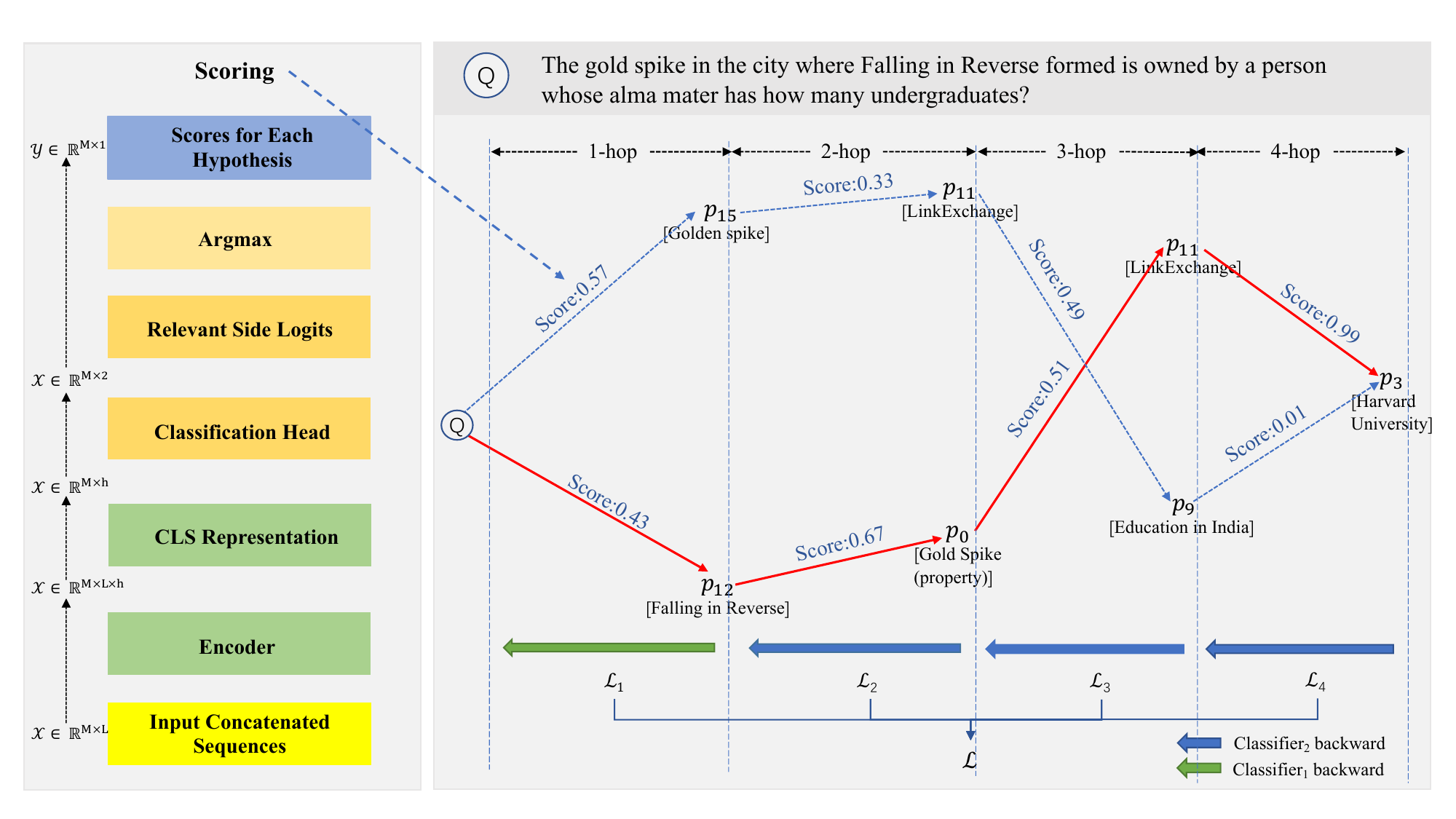}
  \caption{A visualization of Beam Retrieval with a beam size of 2 for the example in Figure~\ref{fig:examples}. The left part shows how to obtain scores for each hypothesis, where M denotes the number of hypotheses at each hop, L denotes the max length of the hypotheses and h denotes the output dimension of the encoder. The right part shows how Beam Retrieval reasons and trains in an end-to-end way, where the red path refers to the ground-truth relevant passages.}
  \label{fig:framework}
\end{figure*}

\section{Beam Retrieval}\label{beam_retrieval_method}

Beam Retrieval is designed to handle a $k$-hop multi-hop question $Q$ and accurately selects the most relevant passages, providing nearly noiseless context for downstream QA tasks. In this section, we clarify how Beam Retrieval infers and trains in an end-to-end manner, which is illustrated in Figure~\ref{fig:framework}.

\subsection{Problem Formulation}\label{score_function}
Given a $k$-hop question $Q$ and a candidate set with $n$ passages as $\mathcal{D} = {\{p_1, p_2, ..., p_n}\}$, multi-hop retrieval aims to produce a relevant passages chain ($\hat{p}_1, \hat{p}_2, ..., \hat{p}_k$). Most existing work formulates it as a one-step or two-step sequence labeling task, classifying every passage $p_i \in \mathcal{D}$ as relevant or not. However, this method lacks generality and precision.

In contrast, we align the multi-hop retrieval task with text decoding, proposing a more general retrieval framework with higher precision. Conceptually, a passage $p_i \in \mathcal{D}$ corresponds to a token $w_i \in \mathcal{V}$ and the question $Q$ corresponds to a special start token ``$<$s$>$''. Similarly, we also denote the output of a multi-hop retriever as $\acute{z}_t = \acute{f}(Q, \hat{p}_1, ..., \hat{p}_{t-1})$, given the concatenated sequence of question and passages identified so far, ($Q, \hat{p}_1, ..., \hat{p}_{t-1}$), which we write as $\hat{p}_{<t}$ for short. The output $\acute{z}_t \in {\mathbb{R}} ^ {n}$. 

We use an auto-encoder language model as an encoder to derive embeddings for the concatenated sequence ($Q, \hat{p}_1, ..., \hat{p}_{t-1}, \acute{z}_t $). Subsequently, a fully connected layer is utilized to project the final dimension of the ``[CLS]'' representations of these embeddings into a 2-dimensional space, representing ``irrelevant'' and ``relevant'' respectively. The logit in the ``relevant'' side serves as the score for the sequence. This scoring process is denoted by a function $S(\acute{z}_t | \hat{p}_{<t})$, and it is shown in Figure~\ref{fig:framework}.

The probability distribution over the next possible relevant passage being $\mathrm{p} \in \mathcal{D}$ is the softmax:

\begin{equation}
\begin{aligned}
\acute{P}(\hat{p}_t = \mathrm{p} | \hat{p}_{<t})&= \frac{S(\acute{z}_{t} | \hat{p}_{<t})} {\sum_{p \in \mathcal{D} \setminus {\{\hat{p}_1, ..., \hat{p}_{t-1}}\}} S(p | \hat{p}_{<t})} \\
&\forall{\acute{z}_{t}} \in \mathcal{D} \setminus {\{\hat{p}_1, ..., \hat{p}_{t-1}}\}
\end{aligned}
\end{equation}

We should keep the uniqueness of each passage within the sequence, as there are no duplicated passages in the only one ground-truth relevant passage chain. This requirement differs from the text decoding process, where such uniqueness is not necessarily enforced.

\subsection{Scoring}
As described in Section~\ref{score_function}, every hypothesis will be scored at each step. Beam Retrieval also employs a scoring function $S(\acute{z}_t | \hat{p}_{<t})$ as illustrated in Figure~\ref{fig:framework}, which utilizes an encoder and two classification heads to obtain scores for each hypothesis of passages. At the first hop, for every passage $p_i \in \mathcal{D}$ we concatenate ``[CLS] + $Q$ + $p_i$ + [SEP]'' to the encoder and derive the encoded $(Q, p_i)$ representations $\textbf{H}^i = [\textbf{h}_1^i, \textbf{h}_2^i, ..., \textbf{h}_{L_i}^i] \in \mathbb{R} ^ {L_i \times h}$, where $L_i$ denotes the length of the concatenated sequence and $h$ denotes the output dimension of the encoder. Then a classification head named ``$classifier_1$'' projects every $\textbf{H}^i$ into a 2-dimensional space, representing  ``irrelevant'' and ``relevant'' respectively. We take the logit in the ``relevant'' side as the score for the sequence $(Q, p_i)$. At subsequent hop $t$, we concatenate ``[CLS] + $Q$ + $\hat{p}_1$ + ... + $\hat{p}_{t-1}$ + $\acute{z}_t$ + [SEP]'' for every $\acute{z}_t \in \mathcal{D} \setminus {\{\hat{p}_1, ..., \hat{p}_{t-1}}\}$. We use the same encoder but another classification head named ``$classifier_2$'' to obtain the score of concatenate sequence $(Q,\hat{p}_1, ..., \hat{p}_{t-1},\acute{z}_t)$ in the same way.  The structures of ``$classifier_1$'' and ``$classifier_2$'' are totally the same, the only difference is ``$classifier_1$'' handles a fixed $n$ sequence while ``$classifier_2$'' deals with a variable number of sequences in an expanded search space.
\subsection{End-to-End Inference}
Compared with previous customized two-step retrieval methods \cite{DBLP:journals/corr/abs-2107-11823, li2023easy, zhangyue-etal-2023-rethinking}, Beam Retrieval employs the beam search paradigm to retrieve multiple relevant passages at each hop, discovering all the relevant passages of $Q$ in an end-to-end way. Let $B$ be the predefined beam size. Starting from the question $Q$, Beam Retrieval pairs it with $n$ passages in $\mathcal{D}$ and scores these $n$ concatenated sequences through the encoder and $classifier_1$, choosing the $B$ passages with the highest scores as the first selected passages. At subsequent hop $t$, Beam Retrieval keeps track of $B$ partial hypotheses, denoted as $\mathcal{P}_{t-1}^b = {\{\hat{p}_1^b, ..., \hat{p}_{t-1}^b}\}$, $b \in [1, B]$. Then we concatenate ($Q$, $\mathcal{P}_{t-1}^b$, $\acute{z}_t$) for every $\acute{z}_t \in \mathcal{D} \setminus {\mathcal{P}_{t-1}^b}$ as input concatenated sequences. In this way Beam Retrieval expands the search space, producing $M$ hypotheses of passages, where $M$ is slightly less than $B \times n$ as we should keep the uniqueness of each passage within the sequence. Then we score these hypotheses using the encoder and $classifier_2$, choosing the $B$ hypotheses with the highest scores. This process continues until the current highest predicted score falls below a predefined threshold $\tau$, and we take the passage sequence from the previous step that has the highest score.

Beam Retrieval finishes the multi-hop retrieval task using a single forward pass, where it calls $k$ times encoder, $1$ time $classifier_1$, and $k-1$ times $classifier_2$. Additionally, as we can see in Figure~\ref{fig:framework}, for methods that select only one passage at a time, choosing an irrelevant passage in the first stage could fail in the entire multi-hop retrieval process.
In conclusion, Beam Retrieval reduces the risk of missing hidden relevant passage sequences by keeping the most likely $B$ hypotheses at each hop and eventually choosing the hypothesis that has the overall highest score.
\subsection{Jointly Optimization}\label{end2end_train}
We jointly optimize the encoder, $classifier_1$, and $classifier_2$ across all hops in an end-to-end manner. Let $(\mathrm{p}_1, \mathrm{p}_2, ..., \mathrm{p}_k)$ be the ground truth relevant passages. At the first hop, the loss can be represented as:

\begin{equation}
\begin{aligned}
\mathcal{L}_{{1}}= & - \sum_{p \in \mathcal{D}} l_{1,p} logS(p | Q) + \\& (1 - l_{1,p})log(1-S(p | Q))
\end{aligned}
\end{equation}
where $l_{1,p}$ is the label of $p$ and $S(p | Q)$ is the score function described in Section~\ref{score_function}. 
At subsequent hop $t$, the loss can be represented as:
\begin{equation}\label{loss_function}
\begin{aligned}
\mathcal{L}_{{t }}= & - \sum_{b=1}^{B} \sum_{p \in \mathcal{D} \setminus {\mathcal{P}_{t-1}^b}} l_{t, p} logS(p | \mathcal{P}_{t-1}^b, Q) \\
& + (1 - l_{t, p})log(1-S(p | \mathcal{P}_{t-1}^b, Q))
\end{aligned}
\end{equation}
where $l_{t, p}$ is the label of $p$. As the beam size $B$ increases, there is a corresponding rise in the number of irrelevant passage sequences. This increment augments Beam Retrieval's capability to accurately identify irrelevant paragraph sequences, allowing the model to halt at the appropriate point during inference, reducing instances of either under-retrieval or over-retrieval of passages.

It is important to note that not all datasets offer the ground-truth relevant passage for each hop. Consequently, for $t \in [1,k]$ we define $l_{t, p}$ under two scenarios: one with a provided order of relevant passages and another without a specified order. If the order of ground-truth relevant passages is given, $l_{t, p}$ is set as: 
\begin{equation}
l_{t, p} = 
\begin{cases}
  1& \text{ if } p= \mathrm{p}_t \\
  0& \text{ if } p \neq \mathrm{p}_t
\end{cases}
\end{equation}

Otherwise $l_{t, p}$ is set as:
\begin{equation}
l_{t, p} = 
\begin{cases}
  1& \text{ if } p \in \{{\mathrm{p}_1, \mathrm{p}_2, ..., \mathrm{p}_k} \}\\
  0& \text{ if } p \notin \{{\mathrm{p}_1, \mathrm{p}_2, ..., \mathrm{p}_k}\}
\end{cases}
\end{equation}

The overall training loss of Beam Retrieval is:
\begin{equation}
\begin{aligned}
\mathcal{L}= \sum_{i=1}^{k} \mathcal{L}_i
\end{aligned}
\end{equation}

\section{Experimental Setup}
\subsection{Datasets}
We focus on the retrieval part of Multi-hop QA and primarily aim at the reading comprehension setting. All experiments are conducted on three benchmark datasets MuSiQue-Ans \cite{musique}, distractor-setting of HotpotQA \cite{yang-etal-2018-hotpotqa} and 2WikiMultihopQA \cite{DBLP:journals/corr/abs-2011-01060}.
For each question, MuSiQue-Ans, HotpotQA, and 2WikiMultihopQA provide 20, 10, and 10 candidate passages, respectively. MuSiQue-Ans requires the model to answer the complicated multi-hop questions, while HotpotQA and 2WikiMultihopQA additionally require the model to provide corresponding supporting sentences. In the setting of Beam Retrieval augmented LLM, we evaluate our method on the partial part of three multi-hop datasets, where we use the 500 questions for each dataset sampled by \cite{trivedi-etal-2023-interleaving}. 

HotpotQA and 2WikiMultihopQA share a similar format and have 2-hop and 2,4-hop questions respectively. Furthermore, 2WikiMultihopQA has entity-relation tuples support, but we do not use this annotation in our training or evaluation. To evaluate Beam Retrieval's performance in more complex scenarios, main experiments are conducted on MuSiQue-Ans, which has 2,3,4-hop questions and is more challenging, as it requires explicit connected reasoning.

\subsection{Models}
\subsubsection{Beam Retrieval}
Beam Retrieval selects all the relevant passages in an end-to-end way. We set the predefined threshold $\tau$ to -1. We employ the base and the large version of DeBERTa \cite{deberta} as our encoder. We use a single RTX4090 GPU and set the number of epochs to 16 and the batch size to 1 (here batch size means the number of examples taken from the dataset, and the actual batch size is the hypothesis number $M$). Owing to our multiple calls of encoder during training, we set gradient checkpointing to True, otherwise it requires a huge amount of memory. We use AdamW \cite{adamw} with a learning rate of 2e-5 for the optimization and set the max position embeddings to 512. Considering the long concatenated sequences, we adopt a truncation method. If the total length exceeds the max length, we calculate the average length of each passage and truncate the extra part. To enhance the robustness of the model, we shuffle the inner order of the concatenated passages within the hypothesis. (See Appendix~\ref{appendix_e} for more details.)

\subsubsection{Downstream Reader}
We implement a downstream reader to receive the retrieved relevant passages as the context $C$, and we concatenate input ``[CLS] + $Q$ + [SEP] + $C$ + [SEP]'' to feed our reader. Specifically, we conduct experiments with two types of readers: supervised setting and few-shot LLM setting.

(i) \textbf{Supervised Reader} 
For MuSiQue-Ans dataset, we train a reading comprehension model following BertForQuestionAnswering \cite{devlin-etal-2019-bert, hugging_face}. For HotpotQA and 2WikiMultihopQA, we train a multi-task reader which extracts the answer and the supporting facts of the question, following FE2H \cite{li2023easy} and R$^3$ \cite{zhangyue-etal-2023-rethinking}, where you can refer to Appendix~\ref{appendix_a} for details. 
In the supervised setting, we employ the large version of DeBERTa for MuSiQue and 2WikiMultihopQA and the xxlarge version of DeBERTa for HotpotQA. We use a single RTX4090 GPU to train the large version reader and a single A100 to train the xxlarge version reader. We set the number of epochs to 12 and the batch size to 4. We use AdamW \cite{adamw} with a learning rate of 5e-6 for the optimization and set the max position embeddings to 1024. To enhance the robustness of the model, we shuffle the inner order of the concatenated passages within the context. (See Appendix~\ref{appendix_e} for more details.)

(ii)\textbf{Few-Shot LLM}
In addition to the supervised reader above, we also incorporate a LLM as the downstream reader to benchmark the few-shot QA performance of Beam Retrieval augmented LLM. In the few-shot LLM setting, given that each example contains up to 20 passages, we choose long-input LLMs. Specifically, we use closed model \emph{gpt-3.5-turbo-16k} provided from API of OpenAI\footnote{\url{https://openai.com/api/}} and open model \emph{longchat-13b-16k}\footnote{\url{https://huggingface.co/lmsys/longchat-13b-16k}} running locally on two 80G-A100 with the help of FastChat\footnote{\url{https://github.com/lm-sys/FastChat}} \cite{fastchat}. We use the template described in Appendix~\ref{appendix_b} to obtain the answers directly.
\subsection{Evaluation Metrics}
Generally, we use Exact Match (EM) and F1 scores to evaluate the retrieval performance. Retrieval EM means whether the passage-level prediction is the same as the ground truth, while retrieval F1 is the harmonic mean of precision and recall, and both of them are irrespective of the inner order between relevant passages. In the retrieve-and-read setting, retrieval EM is particularly critical, as missing relevant passages can significantly impact the performance of downstream readers. 

For MuSiQue-Ans, we report the standard F1-based metrics for the answer (\textbf{An}) and support passage identification (\textbf{Sp}). Actually, \textbf{Sp} F1 in MuSiQue-Ans is equivalent to retrieval F1. For HotpotQA and 2WikiMultihopQA, we report the EM and F1 metrics for the answer prediction task (\textbf{Ans}) and supporting facts prediction task (\textbf{Sup}). In the Beam Retrieval augmented LLM setting, we report the answer F1.

\section{Results}

\paragraph{Influence of Beam Size}
We first explore the influence of different beam sizes on MuSiQue-Ans, as shown in Table~\ref{tab:beam_size}, where the encoder is the base version. Beam Retrieval performs well even with a beam size of 1, showing that modeling the multi-hop retrieval process in an end-to-end manner indeed yields significant improvement, and a beam size of 2 brings further improvement, which is consistent with \cite{seq2seq_learning_with_nn}. However, a beam size greater than 2 leads to a slight decline in performance, which we assume is due to the increase in the number of irrelevant sequences as the beam size expands, making the retrieval task more difficult (further analysis can be found in Appendix~\ref{appendix_c}). It is worth mentioning that in our experimental setting, the candidate set size $n$ ranges from 10 to 20. As the beam size expands, both the necessary training memory and training duration increase rapidly. Due to these considerations, we do not conduct experiments with a beam size larger than 4. In conclusion, we employ beam sizes of 1 and 2 for Beam Retrieval in our subsequent experiments.
\begin{table}[htb]
\centering
\resizebox{\columnwidth}{!}{
\begin{tabular}{ccccc}
\hline
beam size & \textbf{EM}    & \textbf{F1}    & \textbf{Mem} (\%)    & \textbf{Speed} (\%) \\
\hline
1         & 74.18 & 87.46 & 100\% & 100\% \\
2         & \textbf{75.47} & \textbf{88.27} & 119\% & 58\% \\
3         & 74.56 & 87.84 & 150\%   & 42\% \\
4         & 74.43 & 87.65 & 194\%   & 36\% \\
\hline
\end{tabular}
}
\caption{Influence of different beam sizes among retrieval performance, training memory required, and training speed. A beam of size 2 offers the optimal balance between retrieval performance and training costs.}
\label{tab:beam_size}
\end{table}

In terms of computational cost, at each hop Beam Retrieval only calls the encoder and classifier once theoretically, aligning with the resource consumption of previous methods like FE2H and R$^3$, maintaining a similar order of magnitude in both training and inference. We have conducted a specific inference time experiment on the HotpotQA development set to compare the computation cost between Beam Retrieval and past SOTA retrievers R$^3$ and FE2H, where we keep the same backbone model and device. The results are shown in Table~\ref{tab:inference_speed}:

\begin{table}[htb]
\centering
\resizebox{\columnwidth}{!}{
\begin{tabular}{ccccc}
\hline
 & \textbf{Inference Time}    & \textbf{EM}   \\
\hline
FE2H         & \textbf{96.82ms} & 96.35 \\
Smoothing R$^3$         & 127.75ms & 96.85 \\
Beam Retrieval, beam size 1         & 124.64ms & 97.29   \\
Beam Retrieval, beam size 2         & 196.35ms & \textbf{97.52} \\
\hline
\end{tabular}
}
\caption{Comparison of inference speed among Beam Retrieval and past SOTA methods. Beam Retrieval with a beam size of 1 achieves optimal performance while maintaining a similar complexity level as previous methods.}
\label{tab:inference_speed}
\end{table}
Overall, while increasing the beam size does improve performance, it also correspondingly increases computational costs. However, the absolute processing time per question does not become unbearably long. Therefore, we recommend using a beam size of 1 in practical applications, as it offers comparable resource consumption to similar methods, while also achieving superior performance.
\begin{table}[htb]
\centering
\resizebox{\columnwidth}{!}{
\begin{tabular}{cl|cc}
\hline
\multicolumn{2}{c|}{\multirow{2}{*}{\textbf{Methods}}} & \multicolumn{2}{c}{\textbf{Retrieval}} \\ \cline{3-4} 
\multicolumn{2}{c|}{}                             & \textbf{EM}    & \textbf{F1}    \\ \hline
\rowcolor{gray!10} \multicolumn{4}{c}{\textbf{\textsl{MuSiQue-Ans}}} \\ \hline
                    & EE \cite{musique}               
                    & 21.47 & 67.61   \\
                   & SA \cite{musique}               
                    & 30.37 & 72.30         \\
& Ex(EE) \cite{musique}               
                    & 48.78 & 77.79      \\
& Ex(SA) \cite{musique}                
                    & 53.50 & 79.24     \\
& Beam Retrieval, beam size 1 & 77.37 & 89.77       \\ 
                    & Beam Retrieval, beam size 2 & \textbf{79.31} & \textbf{90.51}       \\ 
                    \hline
                    \rowcolor{gray!10} \multicolumn{4}{c}{\textbf{\textsl{HotpotQA}}} \\ \hline
        &SAE \cite{DBLP:conf/aaai/TuHW0HZ20}     & 91.98     & 95.76  \\
        &SA Selector* \cite{musique}               & 93.06 & 96.43         \\
        &S2G \cite{DBLP:journals/corr/abs-2107-11823}  & 95.77     & 97.82 \\
        &FE2H \cite{li2023easy} & 96.32     & 98.02 \\
        &Smoothing R$^3$ \cite{zhangyue-etal-2023-rethinking} & 96.85     & 98.32 \\
        &Beam Retrieval, beam size 1 & 97.29 & 98.55 \\
        &Beam Retrieval, beam size 2 & \textbf{97.52} & \textbf{98.68} \\
        \hline
        \rowcolor{gray!10} \multicolumn{4}{c}{\textbf{\textsl{2WikiMultihopQA}}} \\ \hline
        &SA Selector* \cite{musique}               & 98.25 & 99.13         \\
        &Beam Retrieval, beam size 1 & \textbf{99.93} & \textbf{99.96} \\
                    \hline
\end{tabular}
}
\caption{Retrieval performance on the development set of MuSiQue-Ans, HotpotQA, 2WikiMultihopQA in comparison with previous work. SA Selector* indicates that we reproduce SA Selector by training it on the full HotpotQA and 2WikiMultihopQA. Beam Retrieval surpasses all previous retrievers by a large margin.}
\label{tab:retr_performance}
\end{table}

\begin{table}[htb]
\centering
\resizebox{\columnwidth}{!}{
\begin{tabular}{cl|cc}
\hline
\multicolumn{2}{c|}{\multirow{2}{*}{\textbf{Methods}}} & \multicolumn{2}{c}{\textbf{MuSiQue-Ans}} \\ \cline{3-4} 
\multicolumn{2}{c|}{}                             & \textbf{An}    & \textbf{Sp}    \\ \hline
                    & EE \cite{musique}               
                    & 40.7 & 69.4   \\
                   & SA \cite{musique}               
                    & 52.3 & 75.2         \\
& Ex(EE) \cite{musique}               
                    & 46.4 & 78.1      \\
& Ex(SA) \cite{musique}                
                    & 49.0 & 80.6     \\
& RoHT$^{mix}$ \cite{zhang-etal-2023-reasoning}
                    & 63.6 & 0     \\
& Beam Retrieval, beam size 1 & 66.9 & 90.0       \\ 
                    & Beam Retrieval, beam size 2 & \textbf{69.2} & \textbf{91.4}       \\ 
                    \hline
\end{tabular}
}
\caption{Overall performance on the test set of MuSiQue-Ans. Beam Retrieval achieves a new SOTA.}
\label{tab:total_performance_musique}
\end{table}

\begin{table*}[htb]
\centering
\scalebox{0.9}{
\begin{tabular}{cc|ccccc}
\hline
\multicolumn{2}{c|}{\multirow{2}{*}{\textbf{Methods}}}  & \multicolumn{2}{c}{\textbf{Answer}} & \multicolumn{2}{c}{\textbf{Supporting}} \\ \cline{3-6} 
\multicolumn{2}{c|}{}                             & \textbf{EM}    & \textbf{F1}    & \textbf{EM}    & \textbf{F1}   \\ \hline
\rowcolor{gray!10} \multicolumn{6}{c}{\textbf{\textsl{HotpotQA}}} \\
\hline
                    & HGN \cite{fang-etal-2020-hierarchical}                         &   69.22    &    82.19   & 62.76      &  88.47        \\
                  & SAE \cite{DBLP:conf/aaai/TuHW0HZ20}                         & 66.92      & 79.62      &61.53       &86.86        \\
                   & S2G \cite{DBLP:journals/corr/abs-2107-11823}                         &      70.72 &      83.53 &      64.30 &  88.72      \\
                & FE2H \cite{li2023easy} 
                &      71.89 &  84.44     &  64.98     & 89.14         \\
                    & Smoothing R$^3$ \cite{zhangyue-etal-2023-rethinking}                        &      72.07 &     84.34  &   65.44    & 89.55         \\
                    & Beam Retrieval, beam size 2 & \textbf{72.69} & \textbf{85.04} & \textbf{66.25}    &  \textbf{90.09}       \\ \hline
                    \rowcolor{gray!10} \multicolumn{6}{c}{\textbf{\textsl{2WikiMultihopQA}}} \\ \hline
 & CRERC \cite{CRERC}                      & 69.58 & 72.33 & 82.86  & 90.68   \\
                    & NA-Reviewer \cite{NA-Reviewer}                      &   76.73    & 81.91      &  89.61     &  94.31        \\
                    & BigBird-base model \cite{ho-etal-2023-analyzing}                         &  74.05     &   79.68    &  77.14     &  92.13        \\
                    & Beam Retrieval, beam size 1                        &   \textbf{88.47}    &    \textbf{90.87}   &   \textbf{95.87}    &  \textbf{98.15}        \\
                    \hline
\end{tabular}
}
\caption{Overall performance on the blind test set of HotpotQA and 2WikiMultihopQA in comparison with previous work. Beam Retrieval achieves SOTA in both datasets}
\label{tab:total_performance_HQ_2W}
\end{table*}

\paragraph{Beam Retrieval Performance}
\begin{figure*}[htb]
  \centering
  \includegraphics[width=\linewidth]{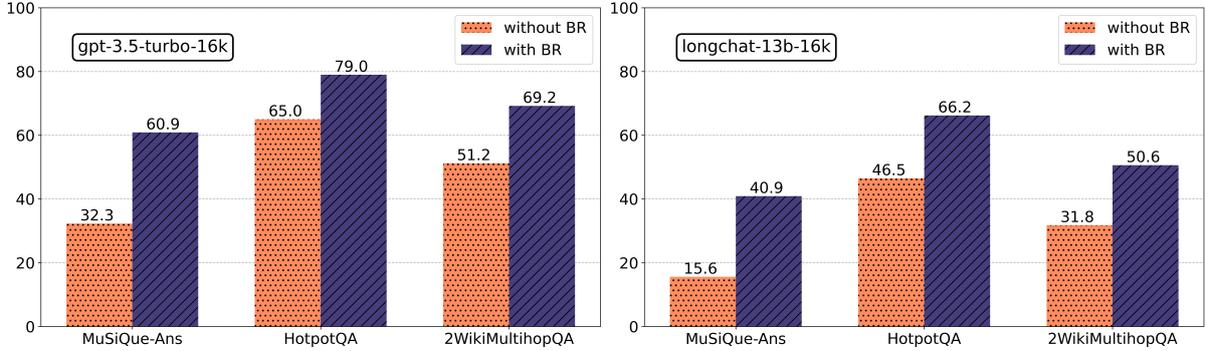}
  \caption{Answer F1 for \emph{gpt-3.5-turbo-16k} (Left) and \emph{longchat-13b-16k} (Right) under two conditions on three multi-hop datasets. Beam Retrieval substantially improves the few-shot QA performance of LLMs.}
  \label{fig:exp_gpt3}
\end{figure*}

We compare our Beam Retrieval with previous retrievers on three multi-hop datasets, as shown in Table~\ref{tab:retr_performance}. Beam Retrieval achieves new SOTA performance across all datasets, significantly outperforming existing methods even when using a beam size of 1, and notably attaining a nearly 50\% EM improvement (from 53.50 to 77.37) on challenging MuSiQue-Ans. This result 
highlights the effectiveness of our proposed end-to-end modeling of the entire multi-hop retrieval process in handling more complex situations. As demonstrated in Table~\ref{tab:beam_size}, employing a beam size of 2 consistently improves performance on both MuSiQue-Ans and HotpotQA datasets, validating the benefits of an expanded search space. As the high-performance retrievers in HotpotQA are customized for two-hop issues, we do not reproduce them for the other two datasets. A large version encoder is employed for all datasets except 2WikiMultihopQA, where a base version encoder achieves a remarkable 99.9\% retrieval precision. Therefore we do not conduct further experiments with larger beam sizes or encoders for this dataset. 

\paragraph{Downstream QA Performance}
Table~\ref{tab:total_performance_musique} and Table~\ref{tab:total_performance_HQ_2W} compare multi-hop QA performance between Beam Retrieval augmented supervised reader (hereinafter referred to as Beam Retrieval) and other strong multi-hop systems across three datasets. Thanks to the retrieved high-quality context, Beam Retrieval with a beam size of 2 achieves new SOTA on all three datasets\footnote{MuSiQue-Ans leaderboard:

\url{https://leaderboard.allenai.org/musique_ans/submissions/public}}\footnote{HotpotQA leaderboard:

\url{https://hotpotqa.github.io/}}\footnote{2WikiMultihopQA leaderboard:\url{https://github.com/Alab-NII/2wikimultihop}}. Specifically, on MuSiQue-Ans our Sp performance (91.4) is comparable to the Human Score (93.9) reported in \cite{musique}. To evaluate the degree of enhancement Beam Retrieval can provide, we compare the few-shot QA performance of few-shot LLMs under two conditions: one using all candidate passages (referred to as ``without BR"), and the other only incorporating relevant passages retrieved by Beam Retrieval (referred to as ``with BR"), which is depicted in Figure~\ref{fig:exp_gpt3}. LLMs perform poorly in directly handling complex multi-hop QA tasks, while Beam Retrieval significantly boosts the few-shot QA performance of both \emph{gpt-3.5-turbo-16k} and \emph{longchat-13b-16k}, some of which are comparable to supervised methods.
\paragraph{Ablation Study}
\begin{table}[htb]
\centering
\resizebox{\columnwidth}{!}{
\begin{tabular}{cl|cc}
\hline
\multicolumn{2}{c|}{\multirow{2}{*}{\textbf{Methods}}} & \multicolumn{2}{c}{\textbf{Retrieval}} \\ \cline{3-4} 
\multicolumn{2}{c|}{}                             & \textbf{EM}    & \textbf{F1}    \\ \hline
& Beam Retrieval$_{1,1}$       & 74.18 & 87.46   \\ 
                    & Beam Retrieval$_{2,2}$       & \textbf{75.47} & \textbf{88.27}   \\ 
                    & Beam Retrieval$_{3,3}$       & 74.56 & 87.84   \\ \hline
                    \rowcolor{gray!10} \multicolumn{4}{c}{\textbf{\textsl{w/o Consistent Beam Size}}} \\ \hline
                   & Beam Retrieval$_{3,2}$       & 74.31 & 87.84   \\
                    & Beam Retrieval$_{3,1}$       & 74.06 & 87.67   \\
                    & Beam Retrieval$_{2,1}$       & 75.13 & 88.17   \\ \hline
                    \rowcolor{gray!10} \multicolumn{4}{c}{\textbf{\textsl{w/o 2 Classification Heads}}} \\ \hline
                    &BR$_{1,1}$ with 4 Classification Heads & 72.16 & 87.04   \\
                    &BR$_{1,1}$ with 1 Classification Head & 73.11 & 87.32   \\
                    \hline
\end{tabular}
}
\caption{Ablation study results on MuSiQue-Ans dataset. The subscript $_{x,y}$ indicates training with beam size $x$ and reasoning with beam size $y$.}
\label{tab:ablation}
\end{table}
To understand the strong performance of Beam Retrieval, we perform an ablation study by employing inconsistent beam sizes between training and reasoning and using different numbers of classification heads, as illustrated in Table~\ref{tab:ablation}. Performance declines when the training beam size differs from the reasoning beam size, and it drops more sharply as the gap between training and reasoning widens. We do not investigate situations where the reasoning beam size exceeds the training beam size, as it is evident that the model cannot perform hard reasoning after easy training. We also vary the number of classification heads to verify if two heads are the optimal setting. First, we use 4 classification heads as there are up to 4-hop questions and we arrange one head for one hop, however it results in a 2-point decrease in EM. Then we employ a unified classification head, which also leads to a one-point performance drop. These results confirm that using one head for the first hop and another head for subsequent hops is the best configuration. We hypothesize that the reason for the 2 heads setup's superior performance is due to the different total number of passages sequences faced at the first hop compared to subsequent hops.
\begin{table}[htb]
\centering
\resizebox{\columnwidth}{!}{
\begin{tabular}{c|c}
\hline
 \textbf{Methods} & \textbf{Retrieval EM}   \\
\hline
MDR (direct) \cite{mdr}         & 65.9 \\
MDR (reranking) \cite{mdr}         & 81.2 \\
MDR (Beam Retrieval reranking) & \textbf{82.2} \\
MDR (gold reranking) & 85.6 \\
\hline
\end{tabular}
}
\caption{Fullwiki HotpotQA reranked retrieval results. Retrieval EM means whether both gold passages are included in the top two retrieved passages (top one chain). Gold reranking refers to whether both gold passages are included among all the retrieved chains.}
\label{tab:fullwiki}
\end{table}

\paragraph{Reranking in Open-Domain Setting}
Beam Retrieval can serve as a reranker in open-domain multi-hop retrieval, and we conduct a simple experiment on fullwiki HotpotQA to assess the impact of Beam Retrieval as a re-ranker, as illustrated in Table~\ref{tab:fullwiki}. We choose MDR \cite{mdr} as the baseline, initially employing it to obtain 100 retrieved passage chains. Subsequently, Beam Retrieval is utilized to rerank the passages within these chains, where we take the top two passages for metric calculation. As an effective reranker, Beam Retrieval further enhances the retrieval performance of open-domain retrieval based on MDR.
\section{Conclusion}
We present Beam Retrieval, an end-to-end beam retrieval framework for multi-hop QA. This approach models the entire retrieval process in an end-to-end manner and maintains multiple partial hypotheses of relevant passages at each step, showing great performance in complex scenarios beyond two hops. Experimental results on three datasets prove the effectiveness of Beam Retrieval and demonstrate it could substantially improve the QA performance of downstream readers. In general, Beam Retrieval establishes a strong baseline for complex multi-hop QA, where we hope that future work could explore more advanced solutions. 

\section*{Limitations}
There are two major limitations to this work. First, the resource consumption during training will increase with larger beam sizes. Second, Beam Retrieval struggles with being independently applied to open-domain settings. We will work on methods to reduce the training consumption of the model and enable its application to open-domain multi-hop retrieval with variable hops.
\section*{Ethics Statement}
This work is a fundamental research work that focuses on technical improvement, thus we have not applied additional filtering techniques to the textual data we used, beyond what has been performed on the original datasets. The textual data we used may have information naming or uniquely identifying individual people or offensive content that we have not been able to identify, as those are out of the focus of this work.
\bibliography{anthology,custom}

\appendix

\section{Multi-Task Supervised Reader}\label{appendix_a}
After receiving the relevant passages ($\hat{p}_1, \hat{p}_2, ..., \hat{p}_k$) from the retriever, our reader is expected to complete both the answer prediction task and the supporting facts prediction task. Following SAE and R$^3$, we also implement a multi-task model to extract the answer and the supporting facts, jointly training the answer prediction and supporting sentence classification in a multi-task learning way.

We define three types of tasks: supporting facts prediction, answer type prediction, and answer span prediction. Following R$^3$, we incorporate a special placeholder token ``$<$d$>$'' before each passage's title and a token ``$<$e$>$'' before each sentence to provide additional information and guide the model to predict at the sentence level.

We concatenate the question and the retrieved passage chain ($\hat{p}_1, \hat{p}_2, ..., \hat{p}_k$) as ``[CLS] + question + [SEP] + $\hat{p}_1$ +  $\hat{p}_2$ + ... + $\hat{p}_k$ + [SEP]''. We denote the BERT-like PLM output as $H=[h1,...,h_L] \in \mathbb{R} ^{L \times d}$ where $L$ is the length of the input sequence and $d$ is the hidden dimension of the backbone model. For answer type prediction, we perform a 3-class ("Yes", "No" and "Span") classification, with the corresponding loss item denoted as  $\mathcal{L}_{type}$. To extract the supporting facts prediction, we apply a linear layer on $H$ to classify each sentence as either a supporting facts sentence or not (using the sentence token ``$<$e$>$''), with its corresponding loss item denoted as $\mathcal{L}_{sf}$. Similarly, we employ another linear layer to project $H$ and identify the start and end positions of the answer, denoting the start position loss and the end position loss as $\mathcal{L}_{start}$ and $\mathcal{L}_{end}$, respectively, as introduced in BERT. Finally, the total answer span loss $\mathcal{L}_{ans}$ is described using the following formulas.

\begin{equation}
\begin{aligned}
\mathcal{L}_{{ans}}= & \lambda_1(\mathcal{L}_{start} + \mathcal{L}_{end})\label{mrc_loss}
\end{aligned}
\end{equation}
where $\lambda_1$ is 0.5 in our setting. Formally, the total loss $\mathcal{L}_{qa}$ can be jointly calculated as:

\begin{equation}
\begin{aligned}
\mathcal{L}_{qa}= & \lambda_2\mathcal{L}_{type} + \lambda_3\mathcal{L}_{sf} + \lambda_4\mathcal{L}_{ans}
\end{aligned}
\end{equation}
where $\lambda_4$ is 0.2 and $\lambda_2, \lambda_3$ are 1 in our setting. Here each loss function is the cross-entropy loss.

\section{Few-Shot Templates}\label{appendix_b}
We use the prompts following \cite{liu2023lost}. To ensure diversity in the demonstrations, we selected demonstrations with different hops and question types. The number of demonstrations is 3.

\begin{prompt}[title={Prompt \thetcbcounter: without Beam Retrieval}]
Write a high-quality answer for the given question using 
only the provided search results (some of which might 
be irrelevant).\\
\\
For example:\\
\\
$\{$examples$\}$\\
\\
$\{$search$\_$results$\}$\\
\\
Question: $\{$question$\}$\\
Answer:
\end{prompt}
\begin{prompt}[title={Prompt \thetcbcounter: with Beam Retrieval}]
Write a high-quality answer for the given question using 
only the provided search results.\\
\\
For example:\\
\\
$\{$examples$\}$\\
\\
$\{$search$\_$results$\}$\\
\\
Question: $\{$question$\}$\\
Answer:
\end{prompt}

\section{Analysis of Beam Search Algorithm in Beam Retrieval}\label{appendix_c}
Beam size is a crucial and interesting parameter in our proposed Beam Retrieval method. It is similar to the beam search process in text decoding, where expanding the search space increases the probability of finding the correct passages. Unlike beam search, Beam Retrieval uses beam size in both the training and inference phases, meaning that the beam size of Beam Retrieval significantly impacts our method's training process. We will dive into the internal workings of Beam Retrieval to explore the actual impact of beam size. As described in the formula~\ref{loss_function}, the essential role of beam size is to increase the number of negative examples in each cross-entropy loss. This enhances the Beam Retrieval's ability to recognize irrelevant passage sequences. As described in the main text, during the inference phase, the decision to stop is based on comparing the highest score of all current passage sequences with a specified threshold $\tau$. Therefore, the ability to identify irrelevant passage sequences determines whether it can stop at an appropriate step. In fact, we conducted experiments on MuSiQue to derive the distribution of scores across different hop counts and beam sizes.

Specifically, we performed an additional hop of reasoning for each question, recording the highest score among all passage sequences at the additional hop.

For instance, for a 4-hop question, we record the score distribution for the abnormal hop 5. All results represent the average scores, and we choose the versions of Beam Retrieval reported in Table~\ref{tab:total_performance_musique} in our original paper, and the results are shown in the table below.
\begin{table}[htb]
\centering
\resizebox{\columnwidth}{!}{
\begin{tabular}{cccc}
\hline
 & \textbf{3hop}    & \textbf{4hop}  & \textbf{5hop} \\
\hline
Additional Hop, beam size 1         & -2.65 & -2.41 &-2.83 \\
Additional Hop, beam size 2         & -2.85 & -2.90 &-3.32 \\
\hline
\end{tabular}
}
\caption{Influence of different beam sizes among the score of the additional hop. It leads to a lower negative score at the additional unreasonable hop as the beam size increases.}
\label{tab:addi_hop}
\end{table}

It is evident that as the beam size increases, it leads to a lower negative score at the unreasonable hop, prompting Beam Retrieval to terminate at the appropriate hop. As the beam size increases, so does the number of negative examples during training, which in turn enhances the model's ability to distinguish irrelevant passage sequences. This allows the model to assign very low negative scores when extraneous passages are introduced, enabling it to terminate in a timely manner. This is the direct cause of the overall performance improvement associated with an increased beam size.

Of course, the beam size can't increase indefinitely. There is indeed a point at which it may lead to suboptimal states. This happens because as the beam size grows, the number of negative instances increases, while there's always only one positive instance. This imbalance makes it increasingly challenging for the model.

Overall, the optimal choice of beam size involves a trade-off that takes into account the size of the backbone model, the number of candidate passages, and the difficulty of the retrieval task. If the backbone model is the base version, the benefits of a larger beam size decrease because an increased beam size expands the search space and brings complexity to the task. Conversely, larger backbone models can get more gains, as demonstrated in the table. The 'large' version of the model achieves a 2 percentage point increase (77.37 -> 79.31) with an increased beam size, while the 'base' version sees only a 1 percentage point improvement (74.18 -> 75.47). The number of candidate passages and the retrieval task difficulty also influence the optimal beam size. For the more challenging MuSiQue task with a larger pool of candidate passages (20), an increase in beam size brings greater benefits. In contrast, for the simpler HotpotQA task with fewer candidate passages (10), the advantages of increasing beam size are less pronounced (97.29 -> 97.52). This is further confirmed by the below experiments on the new knowledge-intensive task IIRC (see Appendix~\ref{appendix_d}), where the questions involve more variable hop counts, thus yielding greater benefits from an increased beam size.

Next, we will discuss why we choose beam search rather than other possible alternatives for multi-hop retrieval. Indeed, there are alternative approaches, such as the coarse-to-fine method proposed by R$^3$. This retriever suggests selecting the three most likely passages in the first hop and then pairing them in the second hop to determine the most probable combination. However, this method has two distinct disadvantages:
\begin{itemize}
  \item Passages not chosen in the first hop are never reconsidered in later stages;
  \item It lacks scalability, as it requires determining too many variables in complex scenarios beyond two hops.
\end{itemize}
Take a three-hop scenario as an example: if you need three passages in the end, how many should you choose in the first hop? (It should be more than three, as there would only be 
 combinations, where n is the number of passages chosen in the first hop, and k is the number of hops. If both n and k equal three, there would only be one candidate.) How many document pairs should be selected in the second hop? And how should a triplet of passages be combined in the third hop from the pairs selected in the second? This process is complicated and involves too many parameters, obviously leading to geometrically increasing computational resource consumption, thus making it highly unscalable.

In contrast, our Beam Retrieval models an end-to-end retrieval process, similar to the decoding of a language model, and naturally incorporates the concept of beam search. Our version with a beam size of 1, where we select only one passage per hop, has outperformed the complex R$^3$ method on HotpotQA dataset. It can also be effortlessly extended to any number of hops, which is verified to adapt Beam Retrieval to another knowledge-intensive task IIRC below, and previous experiments have also validated the effectiveness of Beam Retrieval. We chose beam search over other methods because it was inspired by the striking similarity between multi-hop retrieval process and language model decoding, and it aligns closely with our end-to-end modeling manner.

\section{Beam Retrieval Performance on Dataset IIRC}\label{appendix_d}
As a retrieval method, Beam Retrieval can be adapted for a variety of knowledge-intensive tasks. We have applied Beam Retrieval to another knowledge-intensive dataset IIRC (Incomplete Information Reading Comprehension) \cite{iirc}
  dataset and conducted experiments to evaluate its efficacy. Details and results are given below.

\begin{itemize}
  \item Each question in IIRC is accompanied by an original passage and a set of links to Wikipedia pages which might contain necessary information missing from the original passage. There are a total of 56550 Wikipedia pages and the relevant sentences in pages for each question are given.
  \item We first divide each Wikipedia page into passages consisting of 10 sentences. For each question, we choose the passages with the relevant titles and highest Rouge scores as positive passages and randomly pick passages in irrelevant links as negative passages. Note that for each question, the number of relevant passages varies from 1 to 6 and the number of negative passages varies from 10 to 25, which introduces more uncertainty compared with the three datasets in the paper (This is why we do not choose previous strong retrievers like FE2H and R$^3$ as baselines below). Finally, we get 7566 training samples and 954 test samples.
  \item We train Beam Retrieval (base) on IIRC and compare it with a one-step retriever (like SAE and SA Selector) as the baseline. For Beam Retrieval, nothing changes except concatenating the question and its original passage as the new question text. For baseline, it concatenates each candidate passage and the new question text as inputs fed to BERT, then selects out the most relevant n passages with the highest scores. The retrieval performance is as follows:
\end{itemize}

\begin{table}[htb]
\centering
\resizebox{\columnwidth}{!}{
\begin{tabular}{cc}
\hline
 & \textbf{Retrieval EM}    \\
\hline
One-Step Retriever         & 57.35 \\
Beam Retrieval, beam size 1         & 85.01 \\
Beam Retrieval, beam size 2         & \textbf{86.90} \\
Beam Retrieval, beam size 3         & 86.37 \\
\hline
\end{tabular}
}
\caption{Retrieval performance on IIRC. Beam Retrieval has significantly surpassed the one-step retriever.}
\label{tab:iirc_performance}
\end{table}
We can see a beam size greater than 1 yields better performance. All the results on the new task verify the effectiveness and adaptability of Beam Retrieval.

\section{Ablation Study on Shuffle Operation}\label{appendix_e}
In the training of our Beam Retrieval retriever and supervised reader, we both adopt a shuffle operation to enhance the robustness of the model. 
It is a common and useful deep learning trick that involves shuffling the order of input components. Because we cannot guarantee that passages will always be retrieved in the correct order, we dynamically shuffle the order of input passages during training retriever and reader. To give a specific example, for a 4-hop question, if we have a pair of passages ($p_1, p_2$) by the third hop, we shuffle the order of these two passages. This approach is also applied in subsequent hops, such as the fourth hop, where the order of three passages ($p_1, p_2, p_3$) would be shuffled. Similarly, this shuffling is applied during the training of supervised readers, i.e. shuffle the order of retrieved passages passed to downstream readers. This shuffle operation enhances the model's robustness, allowing it to perform well even if the predicted passages at inference time do not maintain the expected order of reasoning, which is due to that the model was trained with shuffled input sequences. 

We also conducted an ablation study to determine the effectiveness of the shuffle operation on MuSiQue-Ans, as shown in the table below:
\begin{table}[htb]
\centering
\resizebox{\columnwidth}{!}{
\begin{tabular}{ccc}
\hline
\textbf{Methods} & \textbf{An} & \textbf{Sp}   \\
\hline
Beam Retrieval         & 66.9 & 66.9 \\
BR w/o shuffle in retriever         & - & 89.4 \\
BR w/o shuffle in reader         & 64.3 & -\\
\hline
\end{tabular}
}
\caption{Ablation study on shuffle operation.}
\label{tab:shuffle_ablation}
\end{table}
\end{document}